\documentclass{article}

\PassOptionsToPackage{numbers, compress}{natbib}
 \usepackage[preprint]{neurips_2026}

\usepackage[utf8]{inputenc} 
\usepackage[T1]{fontenc}    
\usepackage{hyperref}       
\usepackage{url}            
\usepackage{booktabs}       
\usepackage{amsfonts}       
\usepackage{nicefrac}       
\usepackage{microtype}      
\usepackage{xcolor}         
\usepackage{amsmath}
\newcommand{\panelref}[2]{\hyperref[#1]{\ref*{#1}#2}}
\usepackage{graphicx}

\title{Unifying Dynamical Systems and Graph Theory to Mechanistically Understand Computation in Neural Networks}

\author{%
  Jatin.~Sharma\\
  Department of Electrical and Electronic Engineering\\
  Imperial College London\\
  \And
  Dan FM.~Goodman \\
  Department of Electrical and Electronic Engineering \\
  Imperial College London \\
  \AND
  Danyal.~Akarca \\
  Department of Electrical and Electronic Engineering \\
  Imperial College London \\
}

\begin{document}

\maketitle

\begin{abstract}
Understanding how biological and artificial neural networks implement computation from connectivity is a central problem in neuroscience and machine learning. In neural systems, structural and functional connectivity are known to diverge, motivating approaches that move beyond direct connections alone. Here, we show that the spatial and temporal function of recurrent neural networks (RNNs) trained on hierarchically modular tasks can be recovered by modelling the network as a graph and analysing the multi-hop pathways between input and output units. In particular, decomposing these pathways by hop length reveals how the network temporally routes information. This perspective reframes regularisation: if function is implemented through multi-hop communication, then standard penalties such as L1 regularisation, which act only on individual weights, constrain single-hop structure rather than the multi-hop pathways that support computation. Motivated by this view, we introduce resolvent-RNNs (R-RNNs), which constrain multi-hop pathways and thereby induce temporal sparsity beyond that achieved by standard L1 regularisation. Compared with L1 regularisation, R-RNNs achieve improved performance by inducing temporal sparsity that matches the task structure, even when the task signal is sparse. Moreover, R-RNNs exhibit stronger sparsity-function alignment, reflected in their increased robustness under strong regularisation. Together, our results identify multi-hop communication as a key principle linking structure to function in recurrent networks, and suggest that sparsity should be defined over functional pathways rather than individual parameters. 
\end{abstract}

\section{Introduction}

Computation is often defined as a mapping from inputs to outputs or the manipulation of representations \citep{newell_physical_1980, marr_vision_2010, bengio_representation_2014, saxe_mathematical_2019}. But such definitions say little about how computation is implemented in networks of interacting units evolving through time. This raises a central question in neuroscience and machine learning: can we infer what a network computes from its structure?

A common assumption is that structural connectivity constrains function, such that information flow can be read directly from synaptic or weighted connections. Yet structural and functional connectivity are known to diverge in neural systems \citep{honey_predicting_2009, vazquez-rodriguez_gradients_2019, park_structural_2013}. We argue that one reason is that direct connections capture only single-step interactions, whereas computation in recurrent networks is mediated by multi-hop pathways unfolding over time. Understanding function therefore requires tools that aggregate signal propagation across multiple routes.

Graph theory and dynamical systems provide such a framework. In recurrent networks, temporal evolution can be interpreted as successive hops through connectivity, so graph-based measures such as communicability and the resolvent summarise how structure gives rise to function \citep{zamora-lopez_integrative_2024, estrada_walk-based_2014, fakhar_systematic_2022, crofts_weighted_2009}.

Here we use RNNs as a controlled testbed for this perspective. Across modular tasks requiring averaging, subtraction, addition, and multiplication, we show that the weight matrices alone do not recover the learned input--output structure, even when the network has sufficient capacity to represent it explicitly. By contrast, the resolvent recovers the expected routing structure, and its hop-wise decomposition reveals how information is routed across time as well as space.

Motivated by this result, we ask whether regularisation should target multi-hop pathways rather than individual weights. Standard L1 regularisation acts locally on single connections and therefore does not directly constrain the pathways that support computation. We instead regularise the resolvent, inducing temporally structured sparsity that better aligns routing with task demands and improves performance relative to L1 regularisation.

Together, these results suggest that computation in recurrent networks is better understood through multi-hop pathways than through individual connections. More broadly, this provides a concrete bridge between structure, dynamics, and function in both biological and artificial neural networks (ANNs).

\section{Related Work}

A central challenge in network neuroscience and ML is linking structural connectivity to function \cite{honey_predicting_2009, vazquez-rodriguez_gradients_2019, park_structural_2013}. Increasingly, this is framed as a communication problem: how does information propagate through a network to produce functional interactions \cite{goni_resting-brain_2014, fakhar_systematic_2022}? Shortest-path models are limited because they assume global knowledge of network structure \cite{seguin_brain_2023, achterberg_spatially_2023, seguin_navigation_2018}, motivating walk-based models in which influence is mediated by ensembles of indirect routes.

One such model is communicability, which aggregates walks of all lengths,
\begin{equation}
C = e^A = \sum_{k=0}^{\infty} \frac{A^k}{k!},
\label{C}
\end{equation}
where $A$ is the adjacency matrix \cite{estrada_physics_2012, estrada_walk-based_2014, crofts_weighted_2009}. Related broadcasting models have been shown to capture functionally relevant influence patterns in brain networks \cite{fakhar_general_2024} and have been used to regularise RNNs toward brain-like modular and small-world structure \cite{achterberg_spatially_2023}.

Crucially, these graph measures encode assumptions about dynamics: they are meaningful when the system implements the propagation process they describe \cite{zamora-lopez_integrative_2024}. For example, communicability corresponds to a continuous-time cascade, whereas the resolvent corresponds to a leaky cascade,
\begin{equation}
R = (I - \gamma A)^{-1} = \sum_{k=0}^{\infty}(\gamma A)^k,
\label{R}
\end{equation}
with $\gamma < 1/\lambda_{\max}$ to ensure convergence. This motivates our central question: can such multi-hop measures reconstruct the spatial and temporal structure of computation in ANNs?

\section{Methods}
\label{sec:methods}
\subsection{Task}
\label{sec:task}
We trained RNNs on a family of modular temporal integration tasks designed to probe different input-output routing structures. Inputs consisted of noisy features, $X$, grouped into modules and presented over time, with targets defined as functions of the underlying module-level features. Across tasks, the network had to recover either module means, $\mu$, directly (module averaging as seen in Figure \panelref{all tasks}{a}), combine them linearly through hierarchical subtraction (Figure \panelref{all tasks}{b}) or addition (Figure \panelref{all tasks}{c}), or combine them non-linearly through multiplication (Figure \panelref{all tasks}{d}). Finally, we used a signal on-off task to probe the network's temporal routing structure. Full task definitions are provided in Appendix \ref{sec:Task Definitions} and each task's optimal solution derivation is provided in Appendix \ref{sec:Optimal Solutions Derivation}.

We trained 10 1-layer RNNs on each task where we used a train:validation:test split ratio of $0.64:0.16:0.2$. We used Pytorch's default initialisation on the input-hidden, $W_{ih}\in \mathbb{R}^{i \times h}$, hidden-hidden, $W_{hh}\in \mathbb{R}^{h \times h}$, and hidden-output, $W_{ho}\in \mathbb{R}^{h \times o}$ weights. We applied an L1 loss on the hidden weights, $W_{hh}$, with a regularisation strength of 0.001, except for the multiplication task where no L1 regularisation is applied. We train for 100 epochs on each task using a MSE loss and the Adam optimiser with a learning rate of 0.01. For all our tasks we use $F_{\text{all}}=16$, $M=4$ and $i=h=o=F_{\text{all}}$, $L=5$, and $N=2000$ (defined in Appendix \ref{sec:Task Definitions}). 
\subsection{Computation of Input-Output Multi-Hop Measures}
\label{sec:multi-hop meaures}

We can represent the weight matrices of our RNNs as an adjacency matrix $W$, where $W \in \mathbb{R}^{n \times n}$. Here, $n = i + h + o$, as illustrated in Figure \panelref{networks as graphs}{d}. We aggregate contributions across hop lengths by summing powers of $A$. One convenient way to do so is through the standard definition of the graph resolvent seen in equation \ref{R}. We adapt this by introducing a parameter $\alpha$ such that:

\begin{equation}
    \gamma=\frac{\alpha}{\lambda_{max}}, \quad \text{where} \quad 0<\alpha<1,
\end{equation}
and setting $A=W$ we arrive at:

\begin{equation}
     R = (I - \alpha W^*)^{-1} = \sum_{k=0}^{\infty} (\alpha W^*)^k = I + \alpha W^* + (\alpha W^*)^2 + \cdots ,
\label{R_w_star}
\end{equation}
where $\alpha$ controls the influence of long walk effects and $W^*=\frac{W}{\lambda_{max}}$.
We train RNNs on tasks with a sequence length $L$, so only a finite range of hop lengths can influence the output. We therefore use a truncated resolvent to avoid attributing functional influence to walks longer than those available within the task horizon:
\begin{equation}
     R = \sum_{k=2}^{L+1} (\alpha W^*)^k.
\label{R_w_star_truncated}
\end{equation}
We extract the input-to-output sub-matrix of $R$ to obtain the input-to-output influence map, denoted as $R_{io}$. Specifically, $R_{io} \in \mathbb{R}^{i \times o}$ captures the signed weighted sum over walks between each input and output node, as seen in Figure \panelref{networks as graphs}{d}. For all plots we use $\alpha=0.8$. Similarly, we can compute the input-to-output routing structure associated with an individual hop length by computing $W^k$ and extracting its input-to-output block, giving $W_{io}^k$.

\subsection{Multi-Hop Aware Regularisation: R-RNN}
\label{sec:R-RNN}

In supervised learning, an RNN is trained by adjusting its parameters through backpropagation and gradient descent to minimise a predefined cost function. Typically, this objective combines a task-specific loss, measuring the discrepancy between predicted and true outputs, with an additional weighted sparsity regularisation term, $\beta$, to promote generalisation. The overall loss is given by:
\begin{equation}
    L = L_{\text{task}} + \beta L_{\text{sparsity}}.
\end{equation}
A common choice for $L_{\text{sparsity}}$ is an L1 penalty, where the network is trained to minimise the magnitude of its weights, $L_{\text{sparsity}} = \sum |W|$. In contrast, we train RNNs using a communication based regulariser, where $L_{\text{sparsity}} = \sum |R_{io}|$. Importantly, $R$ is computed from $W$, rather than the normalised $W^*$, to prevent the network from trivially reducing $R_{io}$ by increasing $\lambda_{\text{max}}$. This ensures that reductions in $R_{io}$ reflect a genuine pruning of communication pathways within the network. When $R_{io}$ is used in our regulariser, the truncated sum in equation \ref{R_w_star_truncated} runs from $k=1$ to $L+1$, so that direct one-hop contributions are also penalised. This formulation yields two classes of regularised models: L1-RNNs with $L_{\text{sparsity}} = \sum |W|$, and R-RNNs with $L_{\text{sparsity}} = \sum |R_{io}|$.

We train 10 repeats of each model, for 200 epochs, on the module averaging and oscillating on-off signal tasks. Both models are trained across 10 equally spaced $\beta$ values in the range $[0.0001, 0.01]$. For the R-RNNs we use $\alpha=0.8$. All other task parameters are identical to those described in Section \ref{sec:task}.

\section{Results}

\subsection{Neural Networks as a Graph}

\begin{figure}
    \centering
    \includegraphics[width=1\linewidth]{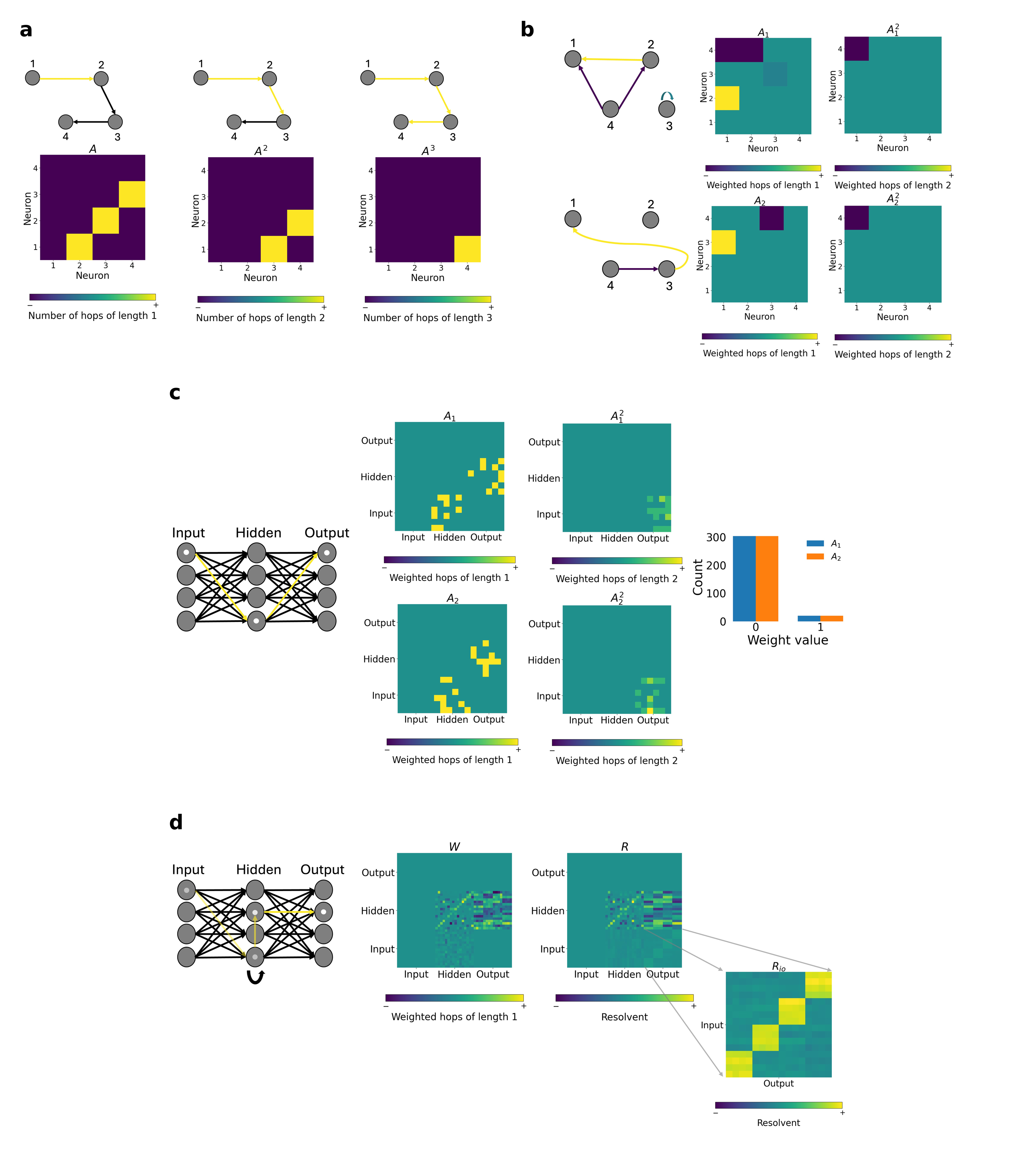}
    \caption{\textbf{Networks represented as graphs and their multi-hop structures.} \textbf{a}, Arbitrary graphs with one pathway highlighted, illustrated by their corresponding $A^k$ matrix. \textbf{b}, Two networks with distinct structures share the same two-hop pattern. \textbf{c}, Two MLPs with the same weight distributions have different two-hop patterns. \textbf{d}, We compute the total input–output influence across multiple hops in an RNN using the resolvent.}
    \label{networks as graphs}
\end{figure}

In graph theory, a network can be represented by nodes and edges, where the structural connectivity is defined by its adjacency matrix, $A$, as seen in Figure \panelref{networks as graphs}{a}. Alternatively, from a dynamical systems perspective, we can view $A$ as representing the number of particle walks of length 1 between each node of the graph. An interesting property of graphs is that $A^k$ tells you how many walks of length $k$ there are between two nodes, as seen in Figure \panelref{networks as graphs}{a}. For example, $A^2_{13}=1$ indicates that there is one possible route of length 2 between nodes 1 and 3. From this we see that a network’s structure ($A$) implicitly defines a family of walk patterns ($A^k$), therefore, if networks transmit information through multiple hops, different network structures give rise to non-trivial differences in network function. For example, Figure \panelref{networks as graphs}{b} shows the $A$ of two structurally different networks; however, both share the same $A^2$ pattern, implying similar function. This observation raises the question of whether representing ANNs as graphs and modelling their multi-hop structure can provide a link between the network's structure and it's function. 

By treating an ANN as a directed graph we can construct its adjacency matrix which contains all possible connections. Figure \panelref{networks as graphs}{c} shows the adjacency matrix for two multi-layer-perceptrons (MLPs), $A_1$ and $A_2$. Both exhibit populated input-hidden and hidden-output blocks, and their weight distributions are identical (\panelref{networks as graphs}{c}). As an MLP has no recurrent connections, a particle can traverse the network only through higher order hops of length two, corresponding to input $\rightarrow$ hidden $\rightarrow$ output paths. To characterise this two hop structure, we compute $A^2$ for each MLP. As expected, only the input–output block is populated. Despite identical weight distributions, the resulting input–output maps differ, reinforcing the non-triviality of hop structure. 

In contrast to the MLP, introducing recurrence through the hidden–hidden connections in our RNN's allows particles to propagate through higher order hops. To capture the cumulative effect of these multi-hop pathways, we calculate the resolvent as defined in the methods and seen in Figure \panelref{networks as graphs}{d}.

\subsection{Multi-Hop Pathways Recover Learned Input-Output Structure but the Weights do not}

\begin{figure}
    \centering
    \includegraphics[width=1\linewidth]{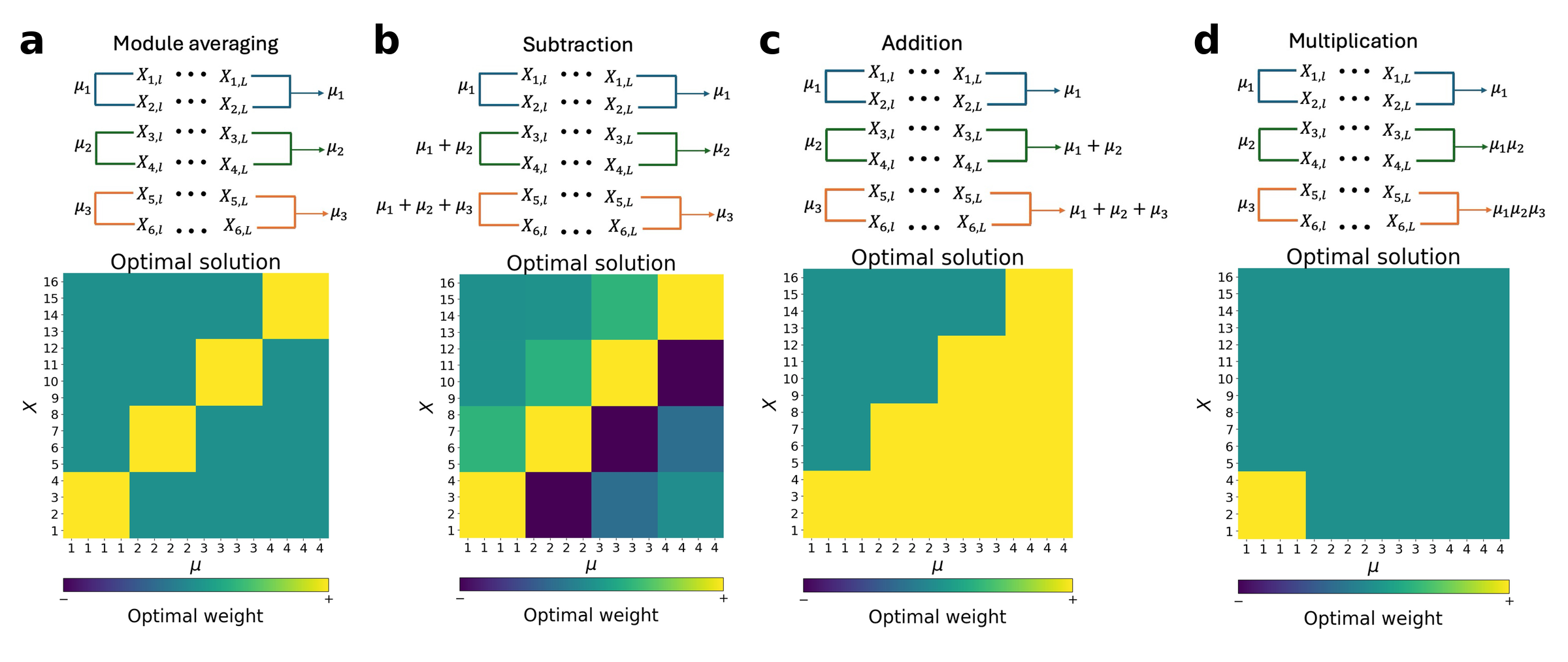}
    \caption{\textbf{Task schematics and their optimal solutions.} \textbf{a}, Module averaging yields strong within-module weights. \textbf{b}, Subtraction yields negative weights between adjacent modules. \textbf{c}, Addition yields hierarchical positive inter-module weights. \textbf{d}, Multiplication yields an input-output map as defined by the task Jacobian.}
    \label{all tasks}
\end{figure}

We aim to show that the RNNs learned solutions can be reconstructed from multi-hop measures. For the task shown in Figure \panelref{all tasks}{a}, we expect the functional routing structure between inputs and outputs to mirror the structure of the optimal solution, as seen in Figure \panelref{all tasks}{a}. This shows for the RNN to achieve the optimal MSE, it must aggregate within module information over time.

\begin{figure}
    \centering
    \includegraphics[width=1\linewidth]{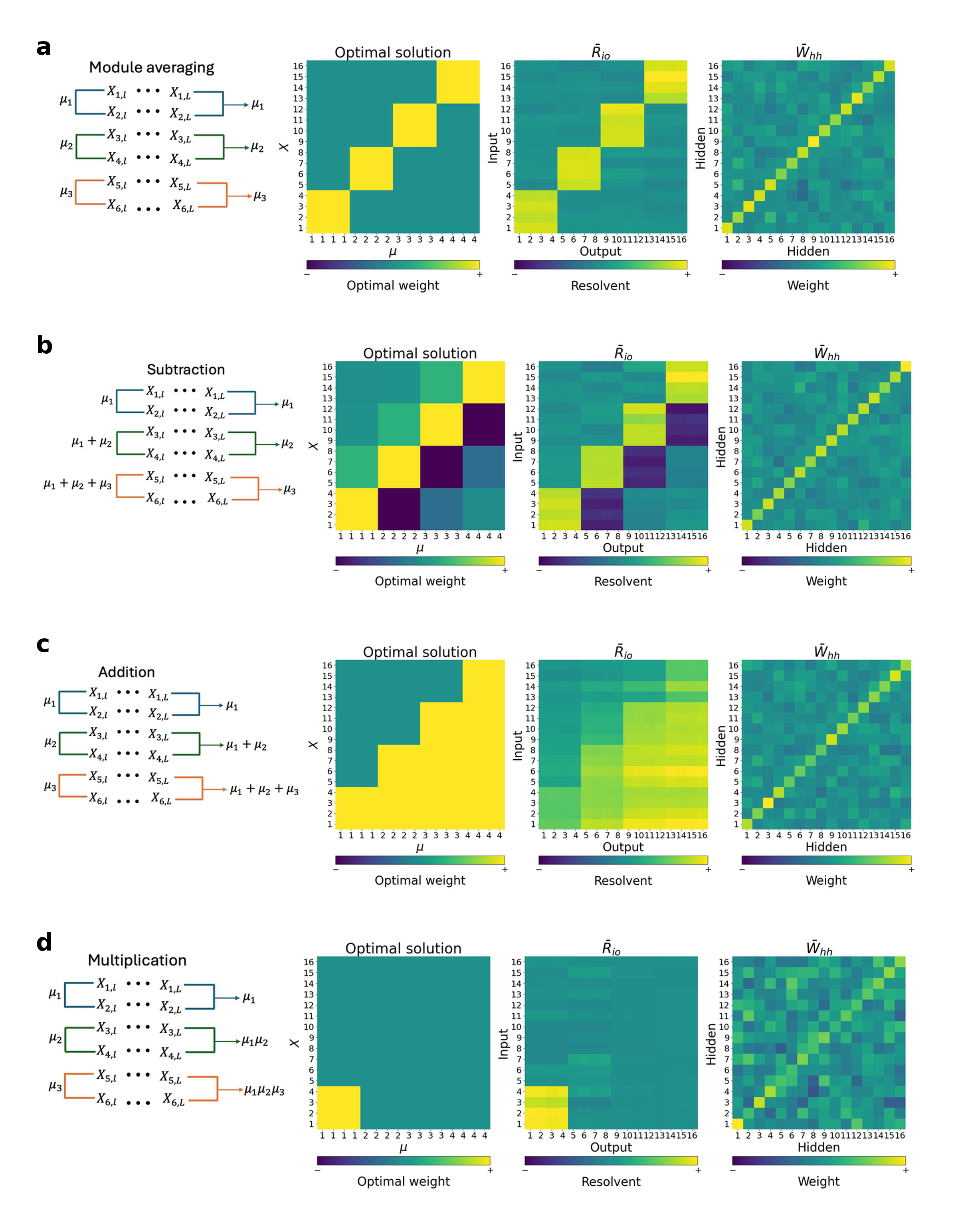}
    \caption{\textbf{The resolvent reconstructs the optimal solution for all tasks but the weights do not.} \textbf{a}, We compare the optimal solution for the module averaging task with the mean $R_{io}$ and $W_{hh}$, and their associated SEM. Correlation with optimal solution (Pearson): $R_{io}$ ($0.9892 \pm 0.0016$) and \ $W_{hh}$ ($0.2383 \pm 0.0193$). \textbf{b}, Analogous comparisons for the subtraction task. Correlation with optimal solution: $R_{io}$ ($0.9903 \pm 0.0003$) and \ $W_{hh}$ ($0.1940 \pm 0.0134$). \textbf{c}, Analogous comparisons for the addition task. Correlation with optimal solution: $R_{io}$ ($0.9351 \pm 0.0105$) and \ $W_{hh}$ ($0.098 \pm 0.0143$). \textbf{d}, Analogous comparisons for the multiplication task. Correlation with optimal solution: $R_{io}$ ($0.9210 \pm 0.0102$) and \ $W_{hh}$ ($0.0305 \pm 0.0221$).}
    \label{R_io and optimal}
\end{figure}

We can gain mechanistic insight into how the RNN computes by viewing information as particles propagating through a directed graph as defined by the RNN's weight matrices. Inputs from each module enter the recurrent state through $W_{ih}$, evolve under the recurrent dynamics $W_{hh}$ for $L$ time steps, and are read out through $W_{ho}$. To ensure the model preserves each module’s information without cross module contamination during recurrent processing, one might expect $W_{hh}$ to be block structured. Because we train networks with $i=h=o=F_{\text{all}}$, the model is not forced to compress representations, so block structure in $W_{hh}$ would be a plausible strategy. However, we do not observe this directly in $W_{hh}$ (Figure \panelref{R_io and optimal}{a}). In contrast, when we account for multi-hop interactions by computing $R_{io}$, the expected module structure emerges clearly (Figure \panelref{R_io and optimal}{a}). These results show that weights alone do not recover the network’s learned input-output structure, whereas multi-hop measures do.

We next ask whether multi-hop measures also explain tasks requiring different operations between modules. To test this, we consider tasks in which information is combined hierarchically across modules (Figure \panelref{all tasks}{}). In the subtraction task, the network must recover individual module means from a cumulative hierarchy, requiring successive subtraction and inhibitory pathways (Figure \panelref{all tasks}{b}). The addition task inverts this structure, requiring successive summation of module means (Figure \panelref{all tasks}{c}). For the product task, the optimal solution is less intuitive because the mapping is non-linear, so we define it using the task Jacobian (Appendix \ref{sec:Optimal Solutions Derivation}).

We clearly see that $R_{io}$ has a higher correlation with the optimal solution for all tasks than $W_{hh}$ (Figure \panelref{R_io and optimal}{}). Small discrepancies are primarily seen in the lowest magnitude entries of $R_{io}$ when compared to the optimal solution. Because trained models achieve near optimal loss, we do not expect $R_{io}$ to reconstruct the optimal solutions exactly for each task. As such, $R_{io}$ offers an interpretable estimate of the information routing learned by the network, recovering the key modular structure even when the match to the analytic optimum is not exact.

\subsection{Hop-Wise Decomposition Reveals Temporal Routing}
\label{sec:Specific Hop Length Contributions}

So far we have shown when the RNN receives signals of the same structure for all time steps, $R_{io}$ reconstructs the spatial structure of the optimal solution. We aim to also understand the temporal routing patterns of the RNN through the lens of hops. The current time agnostic task setup hides the effects individual hop contributions have on the models temporal processing, hence we adapt the module averaging task such that the RNN receives oscillating "on" and "off" signals through time. Specifically, the model receives the same modular signal as seen in Figure \panelref{all tasks}{a} for even time steps ($L$, $L-2$, $L-4$), and then receives standard normally distributed noise at odd time steps ($L-1$, $L-3$). 

How do we expect walks of different lengths to contribute to the RNNs temporal processing for this task? Another way to view this question is as the following: if a signal arrives at the RNN input at time $t$, what walk length $k$ is required for it to influence the network's output? A signal arriving at the input of the one-layer RNN at $t=L-2$ must remain in the hidden state for two time steps before contributing to the output at $t=L$. It therefore follows a walk of length 4:
$W_{ih} \to W_{hh} \to W_{hh} \to W_{ho}$. As such, we expect that hops of length $k$ route inputs arriving at $t=L-(k-2)$. In other words, even numbered walks process signal inputs and odd walks process non-signal inputs. The signal has the same modular structure as the module averaging task, so we expect the network to route signal information in the same pattern as given by the task's optimal solution (Figure \panelref{all tasks}{a}).

\begin{figure}
    \centering
    \includegraphics[width=1\linewidth]{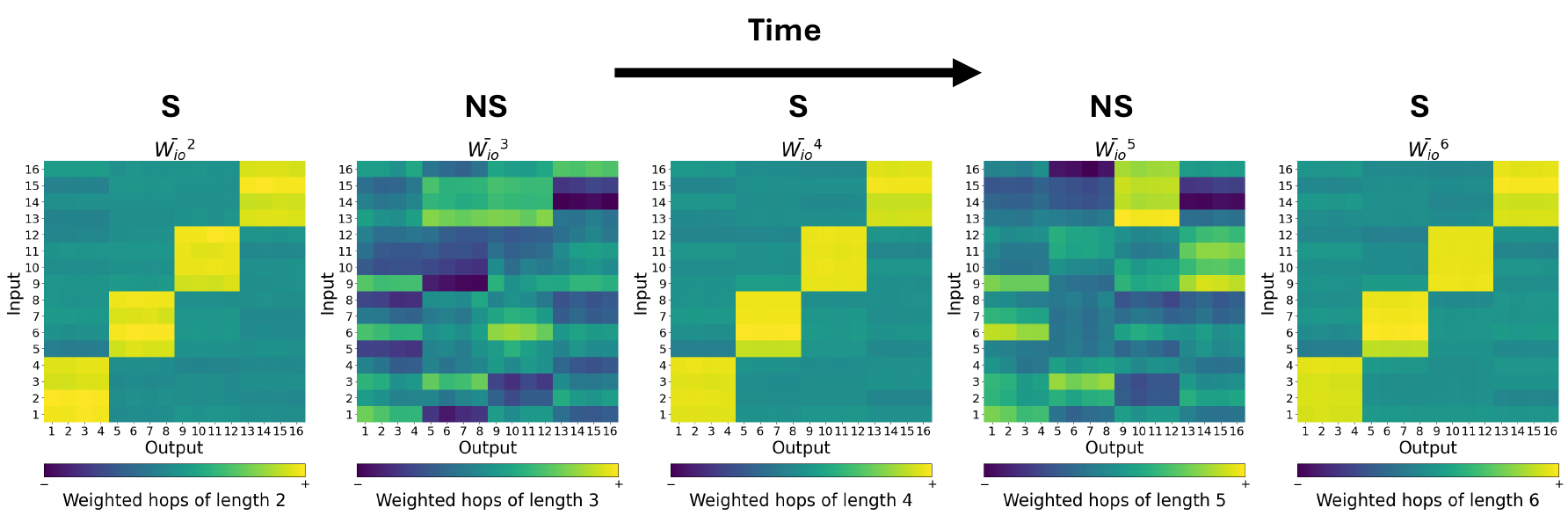}
    \caption{\textbf{$\boldsymbol{W_{io}^k}$ reveals how the network temporally routes information.} The network receives signal (S) and standard normally distributed noise, or no signal (NS), at alternating time steps. As $k$ increases, $W_{io}^k$ corresponds to inputs arriving progressively later in time before the final RNN output.}
    \label{temporal routing}
\end{figure}

Figure \panelref{temporal routing}{} shows the input-output influence for $k=2, 3, 4, 5, 6$. We observe the expected modular structure for even $k$, whereas it disappears for odd $k$. This alternating pattern indicates that $W_{io}^{k}$ faithfully reconstructs the RNN’s expected temporal routing structure. Interestingly, for odd $k$ we see weak block structure that is not task relevant. This is expected, since even length walks must propagate through the recurrent layer to route signal information to the outputs with the correct modular structure. Consequently, both even and odd walks transverse the same structured $W_{hh}$, which can impose a residual block structure even when the timestep carries no functionally useful signal. This highlights a key difference between the Jacobian and hop based measures. The Jacobian measures input dependant sensitivity; for timesteps containing only noise, it can be 0 because the networks outputs are insensitive to those inputs. In contrast, hop based measures reveal the routing pathways through which input signals can propagate. 

\subsection{R-RNNs Outperform L1-RNNs by Encouraging Sparsity Through Time}
\label{sec:R-RNNs vs L1-RNNs}

We now recognise that a network’s spatial and temporal routing patterns can be reconstructed through multi-hop contributions, where the hop order directly encodes the temporal structure of information propagation. This challenges traditional approaches that rely solely on one-hop contributions, as captured by the weights, such as L1 regularisation. 

L1 regularisation encourages network sparsity by penalising the weight magnitude. However, sparse weights do not necessarily imply sparse functional pathways, as a small number of strong connections can still induce dense multi-hop routing. In response to this, we introduce R-RNNs, which directly regularise multi-hop pathways, as outlined in the methods. 

\begin{figure}
    \centering
    \includegraphics[width=1\linewidth]{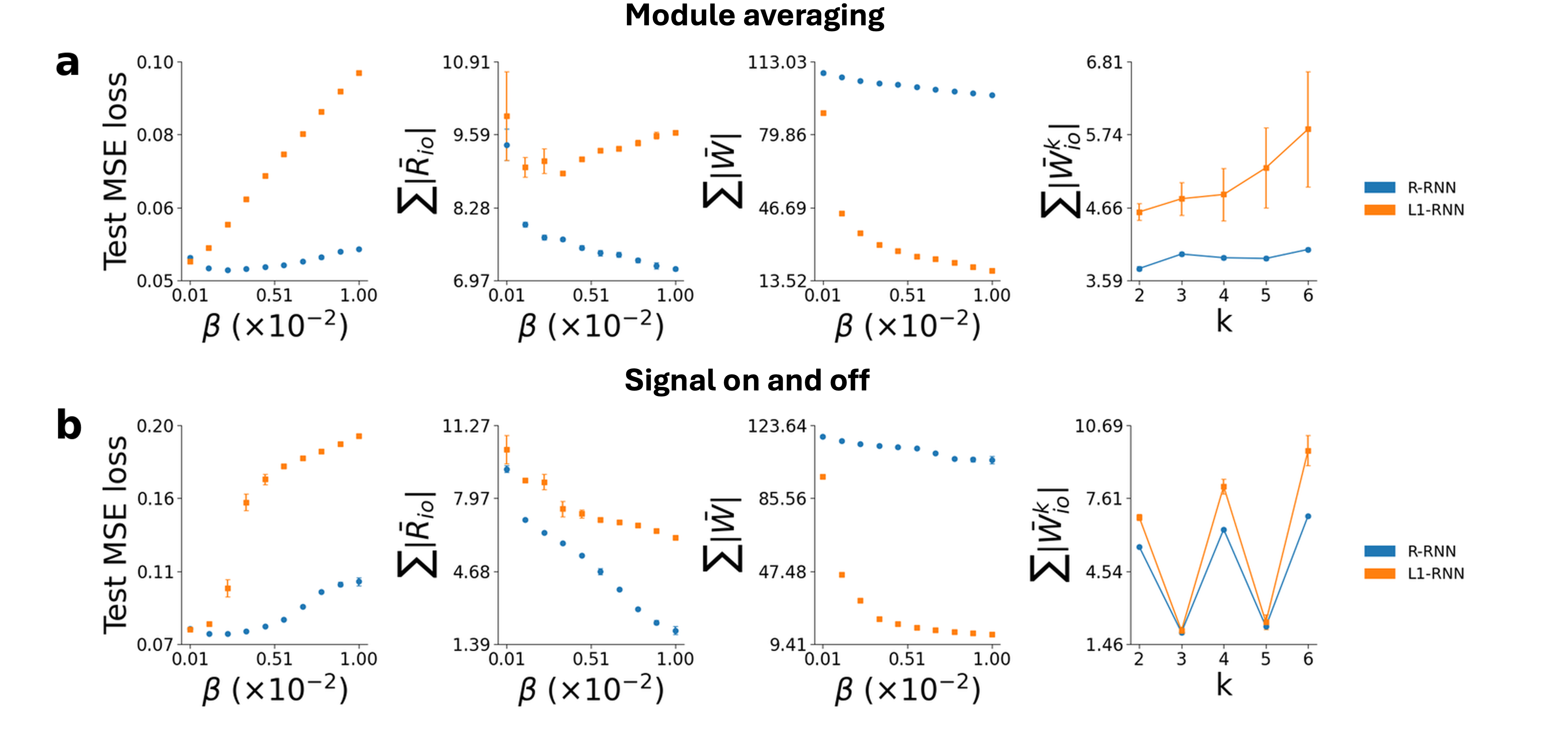}
    \caption{\textbf{R-RNNs outperform L1-RNNs by enforcing sparsity through time.} \textbf{a}, Module averaging task: we compare the test performance and $L_{\text{sparsity}}$ terms of our trained R and L1-RNNs over $\beta$. Additionally, we compare how the total multi-hop magnitude across $k$ varies for the best R-RNN ($\beta = 0.23\times10^{-2}$) and the best L1-RNN ($\beta = 0.01\times10^{-2}$). \textbf{b}, Oscillating on-off signal task: Analogous comparisons. The best R and L1-RNN share the same $\beta$ value as in the module averaging task.}
    \label{R-RNNs vs L1-RNNs}
\end{figure}

Figure \panelref{R-RNNs vs L1-RNNs}{} shows that, although R-RNNs have larger weight magnitudes, they achieve substantially lower resolvent magnitude than L1-RNNs. This reflects a key difference between the two penalties: the resolvent depends on weight placement and sign, not just magnitude, and therefore captures differences in communication pathways that L1 regularisation cannot.

By comparing the test MSE in Figure \panelref{R-RNNs vs L1-RNNs}{}, we observe that, for both tasks, the best R-RNN achieves lower test MSE loss than the best L1-RNN. In Figure \panelref{R-RNNs vs L1-RNNs}{a}, the total hop magnitude of the R-RNN is consistently lower across all $k$. Crucially, the R-RNN equalises hop contributions, matching the task’s routing structure in which the same signal is received at each time step (Figure \panelref{all tasks}{a}). This reflects multi-hop sparsity: the network suppresses redundant pathways across all temporal routes, something L1-RNNs cannot achieve by acting only on individual weights. Additionally, in Figure \panelref{R-RNNs vs L1-RNNs}{b}, the R-RNN selectively suppresses hop communication at $k=2,4,6$, corresponding to signal-carrying time steps (Figure \panelref{temporal routing}{}), while leaving other hops relatively unchanged. This shows that R-RNNs induce temporally structured sparsity that aligns with the task structure even in the presence of sparse signal information 

Furthermore, R-RNNs also perform more robustly under stronger regularisation, indicating that resolvent based sparsity is better aligned with function. As $\beta$ increases, L1-RNNs exhibit a rapid degradation in test performance as total weight magnitude decreases, whereas R-RNNs maintain a small MSE loss even at high levels of resolvent sparsity (Figure \panelref{R-RNNs vs L1-RNNs}{}). This indicates that R-RNNs preserve task relevant computation under strong regularisations. In other words, R-RNNs achieve substantial communication sparsity with only a relatively small loss in performance, demonstrating that multi-hop sparsity is more functionally aligned than weight sparsity.

\section{Discussion}

We  demonstrate that the RNNs input-output mapping is not accounted for by a one hop model as described by $W$, even when $W_{hh}$ is not forced to compress representations. Instead, multi-hop measures recover the spatial-temporal routing structure that mediates information flow from inputs to outputs. This holds across tasks which require different computations: averaging, subtraction, addition, and multiplication, which are likely important for neural information processing. This suggests that multi-hop measures may provide a useful model of communication across a broad range of tasks.

\subsection{Why Multi-Hop Measures Matter for Understanding Computation}

We have shown hop based measures can reveal the input-output routing pathways across both space and time. But why do we care about routing pathways if the Jacobian can tell us functional relationships between the inputs and outputs \cite{pizarroso_neuralsens_2022, yeung_sensitivity_2010, paus_inferring_2005}? While the Jacobian quantifies input-output sensitivity, characterising a biological circuit’s Jacobian across the diversity of naturalistic stimuli is impractical, as it would require extensive controlled experiments. Alternatively, hop based measures can describe a networks "routing capacity" just from its connectivity, therefore providing us with an input agnostic method to characterise which pathways a network can use. As such, we can compare network's "routing capacities" to infer how they would respond to novel stimuli, without requiring new input–output measurements from real neural data.

\subsection{Reconciling Sherringtonian and Hopfieldian Views}

The apparent tension between localised and distributed computation may dissolve under a multi-hop view \cite{barack_two_2021}. Local connectivity (Sherringtonian) can generate distributed routing patterns (Hopfieldian) through the accumulation of walks. What appears as distributed processing reflects the network exploiting many parallel pathways; what appears as localised processing reflects those pathways converging onto specific nodes. The resolvent makes this link explicit, connecting local weights to their global functional consequences.

\subsection{Structure-Function Relationships in Biological Networks}

Importantly, hop measures allow us to connect a network's structure to its function. Biological networks are highly non-random, often demonstrating modular, hierarchical, or small world structures \cite{meunier_modular_2010, sporns_small_2006}. These structural motifs are hypothesised to confer functional benefits like efficient information transfer and flexible learning \cite{gu_emergence_nodate, latora_efficient_2001}. An implicit assumption is that structural modularity is sufficient for functional modularity; however, this assumption has recently been challenged \cite{bena_dynamics_2024}. Multi-hop measures offer a fresh perspective: structural modularity only translates to functional modularity if the walk structure respects module boundaries. Computing $R_{io}$ for networks with varying structural modularity could reveal why the structure-function relationship breaks down in some regimes.

Naturally, these structural motifs assume specific communication patterns, which may also be over represented in biological networks. We can visualise the input-output communication patterns under different connectivity assumptions, which can help us isolate what kinds of computations are supported by structure alone. Additionally, there can exist different connectivity structures with the same $R_{io}$. It would be interesting to test whether structures we typically characterise as distinct nevertheless exhibit similar routing patterns. Moreover, decomposing a networks computation into individual $A^k$ terms, we can characterise different connectivity profiles in terms of their capacity to route information over time.

\subsection{Implications for Mechanistic Interpretability}

A core goal of mechanistic interpretability is producing explanations that humans can understand \cite{bereska_mechanistic_2024}. Graph theoretic descriptions are inherently intuitive as we can represent a networks structure into nodes, edges and particles flowing along paths. Hop based measures decompose a network's computation into contributions from walks of different lengths. This provides a vocabulary for describing what a network does in terms of how information moves through it. Unlike activation based analyses, this approach characterises the network's computational capacity from structure alone.

\subsection{Rethinking Regularisation and Network Design}

If biological networks optimise for efficient communication rather than minimal synaptic strength, then sparsity should be defined over pathways rather than individual connections. In this sense, L1 regularisation can be viewed as a special case of a broader objective: it imposes sparsity at the level of individual weights, but does not constrain how signals propagate through multi-hop pathways. Our results show that this is insufficient for regularising multi-hop communication. R-RNNs achieve this by inducing sparsity in multi-hops, selectively reducing communication to match the task's routing structure. More broadly, the goal of regularisation is to balance task performance with sparsity, and we find R-RNNs better satisfy this trade-off, achieving  communication sparsity with minimal loss in performance. This suggests that functionally aligned sparsity emerges when constraints are placed on how information flows, rather than solely on the underlying parameters.

\section{Limitations and Future Directions}
\label{sec:Limitations and Future Directions}

The resolvent provides a useful multi-hop summary of network communication, but it also assumes a particular dynamical prior: a leaky cascade. Other systems may be better described by alternative measures, such as communicability, which corresponds to a continuous-time cascade and weights walks differently \cite{zamora-lopez_integrative_2024}.

A second limitation is that our tasks are intentionally modular, which raises the question of generality: how well do these measures characterise networks trained on more naturalistic, high-dimensional tasks where structure may be less explicit and representations may be more distributed? Furthermore, do R-RNNs retain their advantages over L1-RNNs in these settings? 

Extending these ideas beyond shallow recurrent models remains an open direction. Applying analogous multi-hop characterisations to deeper or more complex architectures (e.g.Transformers, or mixtures-of-experts) will require careful choices of what constitutes a node and an edge \cite{joshi_transformers_2020, bae_mixture--recursions_2025, muennighoff_olmoe_2025}. 

There may be deeper connections to information theoretic quantities. In particular, do high $R_{io}$ pathways carry more mutual information between inputs and outputs? If so, $R_{io}$ can be used as an additional graph metric for characterising differences in network dynamics between healthy and clinical populations \cite{zhang_mutual_2018, wang_brain_2015}. Multi-hop measures may provide a computationally efficient approximation to input-output structure compared with full input-output sensitivity analysis, which typically depends on repeated evaluations across many inputs

More broadly, the same framework could extend beyond neural networks to other physical computing systems, any setting in which dynamics unfold on a graph and multiple hops shape function \cite{liu_fastgr_2023, barham_pathways_2022}.

\section{Conclusion}

In this work, we show that a network's computation can be viewed through the lens of a graph, where the input-output map can be reconstructed by incorporating multi-hop information, that is not apparent from the raw weight structure alone. This perspective offers a new route for linking structure and function in both biological and artificial networks, and provides a way to probe and compare internal routing strategies for interpretability beyond activation based measures. We show that regularising these multi-hop input–output maps in RNNs improves test performance by inducing temporally structured sparsity. Ultimately, our results point toward a more unified view of computation grounded in structure, dynamics, and the walks that connect them.

{\small
\bibliographystyle{unsrtnat}
\bibliography{references}

@article{liu_fastgr_2023,
	title = {{FastGR}: {Global} {Routing} on {CPU}–{GPU} {With} {Heterogeneous} {Task} {Graph} {Scheduler}},
	volume = {42},
	issn = {1937-4151},
	shorttitle = {{FastGR}},
	url = {https://ieeexplore.ieee.org/abstract/document/9931135},
	doi = {10.1109/TCAD.2022.3217668},
	number = {7},
	urldate = {2026-03-02},
	journal = {IEEE Transactions on Computer-Aided Design of Integrated Circuits and Systems},
	author = {Liu, Siting and Pu, Yuan and Liao, Peiyu and Wu, Hongzhong and Zhang, Rui and Chen, Zhitang and Lv, Wenlong and Lin, Yibo and Yu, Bei},
	month = jul,
	year = {2023},
	pages = {2317--2330},
}

@misc{barham_pathways_2022,
	title = {Pathways: {Asynchronous} {Distributed} {Dataflow} for {ML}},
	shorttitle = {Pathways},
	url = {http://arxiv.org/abs/2203.12533},
	doi = {10.48550/arXiv.2203.12533},
	urldate = {2026-03-02},
	publisher = {arXiv},
	author = {Barham, Paul and Chowdhery, Aakanksha and Dean, Jeff and Ghemawat, Sanjay and Hand, Steven and Hurt, Dan and Isard, Michael and Lim, Hyeontaek and Pang, Ruoming and Roy, Sudip and Saeta, Brennan and Schuh, Parker and Sepassi, Ryan and Shafey, Laurent El and Thekkath, Chandramohan A. and Wu, Yonghui},
	month = mar,
	year = {2022},
	note = {arXiv:2203.12533 [cs]},
}

@article{zhang_mutual_2018,
	title = {Mutual {Information} {Better} {Quantifies} {Brain} {Network} {Architecture} in {Children} with {Epilepsy}},
	volume = {2018},
	copyright = {Copyright © 2018 Wei Zhang et al.},
	issn = {1748-6718},
	url = {https://onlinelibrary.wiley.com/doi/abs/10.1155/2018/6142898},
	doi = {10.1155/2018/6142898},
	language = {en},
	number = {1},
	urldate = {2026-03-02},
	journal = {Computational and Mathematical Methods in Medicine},
	author = {Zhang, Wei and Muravina, Viktoria and Azencott, Robert and Chu, Zili D. and Paldino, Michael J.},
	year = {2018},
	note = {\_eprint: https://onlinelibrary.wiley.com/doi/pdf/10.1155/2018/6142898},
	pages = {6142898},
}

@inproceedings{wang_brain_2015,
	title = {Brain functional connectivity analysis using mutual information},
	url = {https://ieeexplore.ieee.org/document/7418254/},
	doi = {10.1109/GlobalSIP.2015.7418254},
	urldate = {2026-03-02},
	booktitle = {2015 {IEEE} {Global} {Conference} on {Signal} and {Information} {Processing} ({GlobalSIP})},
	author = {Wang, Zhe and Alahmadi, Ahmed and Zhu, David and Li, Tongtong},
	month = dec,
	year = {2015},
	pages = {542--546},
}

@article{paus_inferring_2005,
	title = {Inferring causality in brain images: a perturbation approach},
	volume = {360},
	issn = {0962-8436},
	shorttitle = {Inferring causality in brain images},
	url = {https://doi.org/10.1098/rstb.2005.1652},
	doi = {10.1098/rstb.2005.1652},
	number = {1457},
	urldate = {2026-03-02},
	journal = {Philosophical Transactions of the Royal Society B: Biological Sciences},
	author = {Paus, Tomáš},
	month = may,
	year = {2005},
	pages = {1109--1114},
}

@article{pizarroso_neuralsens_2022,
	title = {{NeuralSens}: {Sensitivity} {Analysis} of {Neural} {Networks}},
	volume = {102},
	copyright = {Copyright (c) 2022 Jaime Pizarroso, José Portela, Antonio Muñoz},
	issn = {1548-7660},
	shorttitle = {{NeuralSens}},
	url = {https://doi.org/10.18637/jss.v102.i07},
	doi = {10.18637/jss.v102.i07},
	language = {en},
	urldate = {2026-03-02},
	journal = {Journal of Statistical Software},
	author = {Pizarroso, Jaime and Portela, José and Muñoz, Antonio},
	month = apr,
	year = {2022},
	pages = {1--36},
}

@book{yeung_sensitivity_2010,
	address = {Berlin, Heidelberg},
	series = {Natural {Computing} {Series}},
	title = {Sensitivity {Analysis} for {Neural} {Networks}},
	copyright = {http://www.springer.com/tdm},
	isbn = {978-3-642-02531-0 978-3-642-02532-7},
	url = {http://link.springer.com/10.1007/978-3-642-02532-7},
	doi = {10.1007/978-3-642-02532-7},
	urldate = {2026-03-02},
	publisher = {Springer},
	author = {Yeung, Daniel S. and Cloete, Ian and Shi, Daming and Ng, Wing W. Y.},
	year = {2010},
}

@article{saxe_mathematical_2019,
	title = {A mathematical theory of semantic development in deep neural networks},
	volume = {116},
	url = {https://www.pnas.org/doi/full/10.1073/pnas.1820226116},
	doi = {10.1073/pnas.1820226116},
	number = {23},
	urldate = {2026-02-19},
	journal = {Proceedings of the National Academy of Sciences},
	publisher = {Proceedings of the National Academy of Sciences},
	author = {Saxe, Andrew M. and McClelland, James L. and Ganguli, Surya},
	month = jun,
	year = {2019},
	pages = {11537--11546},
}

@misc{bengio_representation_2014,
	title = {Representation {Learning}: {A} {Review} and {New} {Perspectives}},
	shorttitle = {Representation {Learning}},
	url = {http://arxiv.org/abs/1206.5538},
	doi = {10.48550/arXiv.1206.5538},
	urldate = {2026-02-19},
	publisher = {arXiv},
	author = {Bengio, Yoshua and Courville, Aaron and Vincent, Pascal},
	month = apr,
	year = {2014},
	note = {arXiv:1206.5538 [cs]},
}

@article{vazquez-rodriguez_gradients_2019,
	title = {Gradients of structure–function tethering across neocortex},
	volume = {116},
	url = {https://www.pnas.org/doi/abs/10.1073/pnas.1903403116},
	doi = {10.1073/pnas.1903403116},
	number = {42},
	urldate = {2026-02-17},
	journal = {Proceedings of the National Academy of Sciences},
	publisher = {Proceedings of the National Academy of Sciences},
	author = {Vázquez-Rodríguez, Bertha and Suárez, Laura E. and Markello, Ross D. and Shafiei, Golia and Paquola, Casey and Hagmann, Patric and van den Heuvel, Martijn P. and Bernhardt, Boris C. and Spreng, R. Nathan and Misic, Bratislav},
	month = oct,
	year = {2019},
	pages = {21219--21227},
}

@article{park_structural_2013,
	title = {Structural and {Functional} {Brain} {Networks}: {From} {Connections} to {Cognition}},
	volume = {342},
	shorttitle = {Structural and {Functional} {Brain} {Networks}},
	url = {https://www.science.org/doi/full/10.1126/science.1238411},
	doi = {10.1126/science.1238411},
	number = {6158},
	urldate = {2026-02-17},
	journal = {Science},
	publisher = {American Association for the Advancement of Science},
	author = {Park, Hae-Jeong and Friston, Karl},
	month = nov,
	year = {2013},
	pages = {1238411},
}

@misc{bereska_mechanistic_2024,
	title = {Mechanistic {Interpretability} for {AI} {Safety} -- {A} {Review}},
	url = {http://arxiv.org/abs/2404.14082},
	doi = {10.48550/arXiv.2404.14082},
	urldate = {2026-02-09},
	publisher = {arXiv},
	author = {Bereska, Leonard and Gavves, Efstratios},
	month = aug,
	year = {2024},
	note = {arXiv:2404.14082 [cs]},
}

@book{marr_vision_2010,
	address = {Cambridge, Mass},
	title = {Vision: a computational investigation into the human representation and processing of visual information},
	isbn = {978-0-262-51462-0 978-0-262-28961-0},
	shorttitle = {Vision},
	language = {en},
	publisher = {MIT Press},
	author = {Marr, David},
	year = {2010},
}

@article{newell_physical_1980,
	title = {Physical symbol systems},
	volume = {4},
	issn = {0364-0213},
	url = {https://www.sciencedirect.com/science/article/pii/S0364021380800152},
	doi = {10.1016/S0364-0213(80)80015-2},
	number = {2},
	urldate = {2026-02-09},
	journal = {Cognitive Science},
	author = {Newell, Allen},
	month = apr,
	year = {1980},
	pages = {135--183},
}

@article{barack_two_2021,
	title = {Two views on the cognitive brain},
	volume = {22},
	copyright = {2021 Springer Nature Limited},
	issn = {1471-0048},
	url = {https://www.nature.com/articles/s41583-021-00448-6},
	doi = {10.1038/s41583-021-00448-6},
	language = {en},
	number = {6},
	urldate = {2026-02-09},
	journal = {Nature Reviews Neuroscience},
	publisher = {Nature Publishing Group},
	author = {Barack, David L. and Krakauer, John W.},
	month = jun,
	year = {2021},
	pages = {359--371},
}

@article{sporns_small_2006,
	title = {Small worlds inside big brains},
	volume = {103},
	url = {https://www.pnas.org/doi/10.1073/pnas.0609523103},
	doi = {10.1073/pnas.0609523103},
	number = {51},
	urldate = {2026-01-29},
	journal = {Proceedings of the National Academy of Sciences},
	publisher = {Proceedings of the National Academy of Sciences},
	author = {Sporns, Olaf and Honey, Christopher J.},
	month = dec,
	year = {2006},
	pages = {19219--19220},
}

@article{meunier_modular_2010,
	title = {Modular and {Hierarchically} {Modular} {Organization} of {Brain} {Networks}},
	volume = {4},
	issn = {1662-4548},
	url = {https://pmc.ncbi.nlm.nih.gov/articles/PMC3000003/},
	doi = {10.3389/fnins.2010.00200},
	urldate = {2025-10-06},
	journal = {Frontiers in Neuroscience},
	author = {Meunier, David and Lambiotte, Renaud and Bullmore, Edward T.},
	month = dec,
	year = {2010},
	pages = {200},
}

@article{honey_predicting_2009,
	title = {Predicting human resting-state functional connectivity from structural connectivity},
	volume = {106},
	url = {https://www.pnas.org/doi/10.1073/pnas.0811168106},
	doi = {10.1073/pnas.0811168106},
	number = {6},
	urldate = {2025-09-18},
	journal = {Proceedings of the National Academy of Sciences},
	publisher = {Proceedings of the National Academy of Sciences},
	author = {Honey, C. J. and Sporns, O. and Cammoun, L. and Gigandet, X. and Thiran, J. P. and Meuli, R. and Hagmann, P.},
	month = feb,
	year = {2009},
	pages = {2035--2040},
}

@misc{bae_mixture--recursions_2025,
	title = {Mixture-of-{Recursions}: {Learning} {Dynamic} {Recursive} {Depths} for {Adaptive} {Token}-{Level} {Computation}},
	shorttitle = {Mixture-of-{Recursions}},
	url = {http://arxiv.org/abs/2507.10524},
	doi = {10.48550/arXiv.2507.10524},
	urldate = {2025-09-16},
	publisher = {arXiv},
	author = {Bae, Sangmin and Kim, Yujin and Bayat, Reza and Kim, Sungnyun and Ha, Jiyoun and Schuster, Tal and Fisch, Adam and Harutyunyan, Hrayr and Ji, Ziwei and Courville, Aaron and Yun, Se-Young},
	month = jul,
	year = {2025},
	note = {arXiv:2507.10524 [cs]},
}

@misc{muennighoff_olmoe_2025,
	title = {{OLMoE}: {Open} {Mixture}-of-{Experts} {Language} {Models}},
	shorttitle = {{OLMoE}},
	url = {http://arxiv.org/abs/2409.02060},
	doi = {10.48550/arXiv.2409.02060},
	urldate = {2025-09-04},
	publisher = {arXiv},
	author = {Muennighoff, Niklas and Soldaini, Luca and Groeneveld, Dirk and Lo, Kyle and Morrison, Jacob and Min, Sewon and Shi, Weijia and Walsh, Pete and Tafjord, Oyvind and Lambert, Nathan and Gu, Yuling and Arora, Shane and Bhagia, Akshita and Schwenk, Dustin and Wadden, David and Wettig, Alexander and Hui, Binyuan and Dettmers, Tim and Kiela, Douwe and Farhadi, Ali and Smith, Noah A. and Koh, Pang Wei and Singh, Amanpreet and Hajishirzi, Hannaneh},
	month = mar,
	year = {2025},
	note = {arXiv:2409.02060 [cs]},
}

@misc{joshi_transformers_2020,
	title = {Transformers are {Graph} {Neural} {Networks}},
	url = {https://graphdeeplearning.github.io/post/transformers-are-gnns/},
	language = {en-us},
	urldate = {2025-08-26},
	journal = {NTU Graph Deep Learning Lab},
	author = {Joshi, Chaitanya},
	month = feb,
	year = {2020},
}

@article{estrada_physics_2012,
	title = {The {Physics} of {Communicability} in {Complex} {Networks}},
	volume = {514},
	issn = {0370-1573},
	url = {http://arxiv.org/abs/1109.2950},
	doi = {10.1016/j.physrep.2012.01.006},
	number = {3},
	urldate = {2025-07-22},
	journal = {Physics Reports},
	author = {Estrada, Ernesto and Hatano, Naomichi and Benzi, Michele},
	month = may,
	year = {2012},
	note = {arXiv:1109.2950 [physics]},
	pages = {89--119},
}

@article{goni_resting-brain_2014,
	title = {Resting-brain functional connectivity predicted by analytic measures of network communication},
	volume = {111},
	url = {https://www.pnas.org/doi/abs/10.1073/pnas.1315529111},
	doi = {10.1073/pnas.1315529111},
	number = {2},
	urldate = {2025-07-14},
	journal = {Proceedings of the National Academy of Sciences},
	publisher = {Proceedings of the National Academy of Sciences},
	author = {Goñi, Joaquín and van den Heuvel, Martijn P. and Avena-Koenigsberger, Andrea and Velez de Mendizabal, Nieves and Betzel, Richard F. and Griffa, Alessandra and Hagmann, Patric and Corominas-Murtra, Bernat and Thiran, Jean-Philippe and Sporns, Olaf},
	month = jan,
	year = {2014},
	pages = {833--838},
}

@article{seguin_navigation_2018,
	title = {Navigation of brain networks},
	volume = {115},
	copyright = {Copyright © 2018 the Author(s). Published by PNAS.},
	url = {https://www.pnas.org/doi/abs/10.1073/pnas.1801351115},
	doi = {10.1073/pnas.1801351115},
	language = {EN},
	number = {24},
	urldate = {2025-07-08},
	journal = {Proceedings of the National Academy of Sciences},
	publisher = {Proceedings of the National Academy of Sciences},
	author = {Seguin, Caio and van den Heuvel, Martijn P. and Zalesky, Andrew},
	month = jun,
	year = {2018},
	note = {Company: National Academy of Sciences
Distributor: National Academy of Sciences
Institution: National Academy of Sciences
Label: National Academy of Sciences},
	pages = {6297--6302},
}

@article{estrada_walk-based_2014,
	title = {Walk-based measure of balance in signed networks: {Detecting} lack of balance in social networks},
	volume = {90},
	shorttitle = {Walk-based measure of balance in signed networks},
	url = {https://link.aps.org/doi/10.1103/PhysRevE.90.042802},
	doi = {10.1103/PhysRevE.90.042802},
	number = {4},
	urldate = {2025-07-04},
	journal = {Physical Review E},
	publisher = {American Physical Society},
	author = {Estrada, Ernesto and Benzi, Michele},
	month = oct,
	year = {2014},
	pages = {042802},
}

@article{gu_emergence_nodate,
	title = {Emergence and reconfiguration of modular structure for artificial neural networks during continual familiarity detection},
	volume = {10},
	issn = {2375-2548},
	url = {https://www.ncbi.nlm.nih.gov/pmc/articles/PMC11277393/},
	doi = {10.1126/sciadv.adm8430},
	number = {30},
	urldate = {2025-06-17},
	journal = {Science Advances},
	author = {Gu, Shi and Mattar, Marcelo G. and Tang, Huajin and Pan, Gang},
	pages = {eadm8430},
}

@article{latora_efficient_2001,
	title = {Efficient {Behavior} of {Small}-{World} {Networks}},
	volume = {87},
	copyright = {http://link.aps.org/licenses/aps-default-license},
	issn = {0031-9007, 1079-7114},
	url = {https://link.aps.org/doi/10.1103/PhysRevLett.87.198701},
	doi = {10.1103/PhysRevLett.87.198701},
	language = {en},
	number = {19},
	urldate = {2025-06-16},
	journal = {Physical Review Letters},
	author = {Latora, Vito and Marchiori, Massimo},
	month = oct,
	year = {2001},
	pages = {198701},
}

@article{fakhar_systematic_2022,
	title = {Systematic perturbation of an artificial neural network: {A} step towards quantifying causal contributions in the brain},
	volume = {18},
	issn = {1553-7358},
	shorttitle = {Systematic perturbation of an artificial neural network},
	url = {https://journals.plos.org/ploscompbiol/article?id=10.1371/journal.pcbi.1010250},
	doi = {10.1371/journal.pcbi.1010250},
	language = {en},
	number = {6},
	urldate = {2024-11-26},
	journal = {PLOS Computational Biology},
	publisher = {Public Library of Science},
	author = {Fakhar, Kayson and Hilgetag, Claus C.},
	month = jun,
	year = {2022},
	pages = {e1010250},
}

@misc{bena_dynamics_2024,
	title = {Dynamics of specialization in neural modules under resource constraints},
	url = {http://arxiv.org/abs/2106.02626},
	urldate = {2024-11-12},
	publisher = {arXiv},
	author = {Béna, Gabriel and Goodman, Dan F. M.},
	month = oct,
	year = {2024},
	note = {arXiv:2106.02626},
}

@article{seguin_brain_2023,
	title = {Brain network communication: concepts, models and applications},
	volume = {24},
	copyright = {2023 Springer Nature Limited},
	issn = {1471-0048},
	shorttitle = {Brain network communication},
	url = {https://www.nature.com/articles/s41583-023-00718-5},
	doi = {10.1038/s41583-023-00718-5},
	language = {en},
	number = {9},
	urldate = {2024-11-12},
	journal = {Nature Reviews Neuroscience},
	publisher = {Nature Publishing Group},
	author = {Seguin, Caio and Sporns, Olaf and Zalesky, Andrew},
	month = sep,
	year = {2023},
	pages = {557--574},
}

@misc{zamora-lopez_integrative_2024,
	title = {An integrative dynamical perspective for graph theory and the study of complex networks},
	url = {http://arxiv.org/abs/2307.02449},
	doi = {10.48550/arXiv.2307.02449},
	urldate = {2024-11-12},
	publisher = {arXiv},
	author = {Zamora-López, Gorka and Gilson, Matthieu},
	month = feb,
	year = {2024},
	note = {arXiv:2307.02449},
}

@book{fakhar_general_2024,
	title = {A {General} {Framework} for {Characterizing} {Optimal} {Communication} in {Brain} {Networks}},
	doi = {10.1101/2024.06.12.598676},
	author = {Fakhar, Kayson and Hadaeghi, Fatemeh and Seguin, Caio and Dixit, Shrey and Messé, Arnaud and Zamora-López, Gorka and Misic, Bratislav and Hilgetag, Claus},
	month = jun,
	year = {2024},
}

@article{crofts_weighted_2009,
	title = {A weighted communicability measure applied to complex brain networks},
	volume = {6},
	issn = {1742-5689},
	doi = {10.1098/rsif.2008.0484},
	language = {eng},
	number = {33},
	journal = {Journal of the Royal Society, Interface},
	author = {Crofts, Jonathan J. and Higham, Desmond J.},
	month = apr,
	year = {2009},
	pages = {411--414},
}

@article{achterberg_spatially_2023,
	title = {Spatially embedded recurrent neural networks reveal widespread links between structural and functional neuroscience findings},
	volume = {5},
	copyright = {2023 The Author(s)},
	issn = {2522-5839},
	url = {https://www.nature.com/articles/s42256-023-00748-9},
	doi = {10.1038/s42256-023-00748-9},
	language = {en},
	number = {12},
	urldate = {2024-11-12},
	journal = {Nature Machine Intelligence},
	publisher = {Nature Publishing Group},
	author = {Achterberg, Jascha and Akarca, Danyal and Strouse, D. J. and Duncan, John and Astle, Duncan E.},
	month = dec,
	year = {2023},
	pages = {1369--1381},
}
}


\appendix

\section{Appendix}
\subsection{Task Definitions}
\label{sec:Task Definitions}
Our RNNs are initially trained on a task that requires them to average noisy modular features over time. We first define the task's hidden parameters given by:

\begin{equation}
    \mu_h \sim \mathcal{N}(0, \sigma_\mu^2),
    \label{hidden params}
\end{equation}
and then the module parameters

\begin{equation}
    \mu_m = G_{hm}\mu_h,
\end{equation}
where $G_{hm}$ defines the module structure. In our tasks we have two module structures given by

\begin{equation}
    G^{\text{mod}}_{hm}= \begin{pmatrix}
                        1 & 0 & 0\\
                        0 & 1 & 0\\
                        0 & 0 & 1
                        \end{pmatrix}, \quad \text{and} \quad
     G^{\text{add}}_{hm}= \begin{pmatrix}
                        1 & 0 & 0\\
                        1 & 1 & 0\\
                        1 & 1 & 1
                        \end{pmatrix},
\end{equation}
which are shown for tasks with $M$ = 3 modules. Next, we define our feature parameters as

\begin{equation}
    \mu_f = G_{mf}\mu_m,
\end{equation}
and $G_{mf}=I_M \otimes \mathbf{1}_F$, where $F$ is the number of features per module. Given the feature parameters, we then take normally distributed noisy samples given by:

\begin{equation}
    X \sim \mathcal{N}(\mu_f, \sigma_\epsilon^2I_F),
\end{equation}
and $N$ independent samples are taken over $L$ time steps, such that $X \in \mathbb{R}^{N \times L \times F_{\text{all}}}$, where $F_{\text{all}} = F M$ denotes the total number of features. For the signal on-off task used in Sections \ref{sec:Specific Hop Length Contributions} and \ref{sec:R-RNNs vs L1-RNNs}, standard normally distributed noise is injected at odd time steps (i.e.\ $L-1$, $L-3$, \dots). For each task, the network receives $X$ as input and must predict the following target parameters:

\begin{equation}
    \mu_t=f_t(\mu_m),
\end{equation}
where $f_t$ is some function that maps our task parameters to the network's target parameters.  

By using different choices of $G_{hm}$ and $f_t$, we can define our set of tasks as shown in Figure \panelref{all tasks}{}. First, we define the modular averaging task, which uses $G^{\text{mod}}_{hm}$ and $f_t(\mu_m) = \mu_m$ (Figure \panelref{all tasks}{a}). Next, we define the subtraction and addition tasks. The subtraction task uses $G^{\text{add}}_{hm}$ with $f_t(\mu_m) = A^{\text{sub}}\mu_m$ (Figure \panelref{all tasks}{b}), while the addition task uses $G^{\text{mod}}_{hm}$ with $f_t(\mu_m) = A^{\text{add}}\mu_m$. We define $A^{\text{sub}}$ and $A^{\text{add}}$ as:

\begin{equation}
\begin{aligned}
    A^{\text{sub}} &= 
    \begin{pmatrix}
        1 & 0 & 0\\
        -1 & 1 & 0\\
        0 & -1 & 1
    \end{pmatrix}
    \quad \text{and} \quad
    A^{\text{add}} &= 
    \begin{pmatrix}
        1 & 0 & 0\\
        1 & 1 & 0\\
        1 & 1 & 1
    \end{pmatrix}
\end{aligned}
\end{equation}
Finally, we define the multiplication task, which uses $G^{\text{mod}}_{hm}$ with $f_t(\mu_m)$ as:
\begin{equation}
    \begin{pmatrix}
        \mu_1 \\
        \mu_1 \mu_2 \\
        \mu_1 \mu_2 \mu_3
    \end{pmatrix},
\end{equation}
as shown in Figure \panelref{all tasks}{d}. Across all tasks, the module level targets are shared among features within the same module, and $N$ independent samples are drawn to form $\mu_t \in \mathbb{R}^{N \times F_{\text{all}}}$.
\subsection{Optimal Solutions Derivation}
\label{sec:Optimal Solutions Derivation}
In this section, we will discuss how we arrived at the optimal solutions as seen in Figure \ref{all tasks}. We can represent our module averaging, subtraction and addition tasks as a linear regression problem:

\begin{equation}
    X = A\mu + \epsilon,
    \label{regression}
\end{equation}
where $X \in \mathbb{R}^{n}$, such that $n=F_{all} \times L$, defines our feature vector for all modules over time, $A \in \mathbb{R}^{n \times M}$ defines our task structure, $\mu \in \mathbb{R}^{M}$ defines our module means, and $\epsilon \in \mathbb{R}^{n}$ defines the feature noise. Using the standard result of the posterior mean under Bayesian linear regression we obtain:
\begin{equation}
    \hat{\mu}=W_{\text{optimal}}X, \quad \text{where} \quad    W_{\text{optimal}} = (A^\top A + \frac{\sigma_\epsilon^2} {\sigma_\mu^2} I)^{-1} A^\top,
\end{equation}
such that $W_{\text{optimal}} \in \mathbb{R}^{M \times n}$. We then reshape such that $W_{\text{optimal}} \in \mathbb{R}^{M \times F_{\text{all}} \times L}$, average over the temporal dimension $L$, and duplicate the resulting module level solutions. This produces a final matrix $W_{\text{optimal}} \in \mathbb{R}^{F_{\text{all}} \times F_{\text{all}}}$, as shown in Figure \ref{all tasks}.

For the multiplication task (Figure \panelref{all tasks}{d}) we define the module level task structure with input vector $I$, where $I = (\mu_1, \mu_2, \ldots, \mu_M)^\top$, and output vector $O$, where $O = (\mu_1, \mu_1\mu_2, \ldots, \prod_{k=1}^{M} \mu_k)^\top$. The inputs cannot be represented as a linear combination of the network outputs, as governed by equation \ref{regression}. Instead, we calculate the input-output sensitivity by computing the task's Jacobian, $J$, where each element is given by:

\begin{equation}
    J_{ij} = \frac{\partial O_j}{\partial I_i}.
\end{equation}
Since $O_j = \prod_{k=1}^{j} \mu_k$, we obtain:

\begin{equation}
    J_{ij}
    =
    \begin{cases}
        \dfrac{O_j}{\mu_i}, & i \leq j, \\
        0, & i > j.
    \end{cases}
\end{equation}
In practice, the module level means are sampled from normal distributions as shown in equation \ref{hidden params}, and given that the network is trained on $N$ independent samples, we have $\mathbb{E}[\mu_i] = 0$. For example, when $M=3$, the Jacobian displayed in reverse row order takes the form
\begin{equation}
J =
\begin{pmatrix}
0 & 0 & \mu_1\mu_2 \\
0 & \mu_1 & \mu_1\mu_3 \\
1 & \mu_2 & \mu_2\mu_3
\end{pmatrix},
\end{equation}
where, in expectation, all multiplicative terms vanish, leaving only $J_{1,1}$ populated. By duplicating this module level Jacobian structure across features, we obtain the corresponding optimal solution, as shown in Figure \panelref{all tasks}{d}.

\subsection{Implementation Details}
\label{sec:Implementation Details}

Most implementation details are provided in Section \ref{sec:methods}; here we include additional information. Unless otherwise stated, all reported uncertainties and error bars denote $\pm 1$ standard error of the mean (SEM) across 10 independent runs with different random initialisations. All models were trained locally on CPU using an Apple M1 MacBook Pro. Training 10 repeats for 100 epochs, as described in Section \ref{sec:task}, took roughly 15 s. Training R-RNN and L1-RNN models for 200 epochs across $\beta$ values, as described in Section \ref{sec:R-RNN}, took roughly 2 mins.

\newpage

\end{document}